\documentclass[10pt,twocolumn,letterpaper]{article}

\usepackage[pagenumbers]{iccv}              

\usepackage{multirow}
\usepackage{placeins} 
\usepackage{dblfloatfix}
\usepackage[accsupp]{axessibility}
\usepackage{cuted}

\definecolor{iccvblue}{rgb}{0.21,0.49,0.74}
\usepackage[pagebackref,breaklinks,colorlinks,allcolors=iccvblue]{hyperref}

\def\paperID{6636} 
\def\confName{ICCV}
\def\confYear{2025}

\NewDocumentCommand{\var}{O{s} m O{}}{%
  \ensuremath{#1_{#2}^{#3}}
}
\usepackage{siunitx}

\newcommand{\commentout}[1]{}

\definecolor{light-gray}{gray}{0.80}

\newcommand\fref{Fig.~\ref}
\newcommand\tref{Tab.~\ref}

\usepackage{amsthm}

\usepackage{amsthm}

\newtheorem{mytheorem}{Theorem}
\newtheorem{assumption}[mytheorem]{Assumption}
\newtheorem{lemma}[mytheorem]{Lemma}
\newtheorem{remk}[mytheorem]{Remark}

\newcommand{\name}{InstantEdit\xspace}

\definecolor{darkred}{rgb}{0.7, 0, 0}
\definecolor{darkblue}{rgb}{0, 0, 0.9}

\title{InstantEdit: Text-Guided Few-Step Image Editing with Piecewise Rectified Flow}

\author{
Yiming Gong, Zhen Zhu, Minjia Zhang \\ [4mm] 
University of Illinois at Urbana-Champaign \\ 
{\tt\small \{yimingg8, zhenzhu4, minjiaz\}@illinois.edu} 
}

\begin{document}
\maketitle

\begin{strip}
\centering
\includegraphics[width=\textwidth]{figs/supp_diverse_edit.pdf}
\captionof{figure}{Examples of InstantEdit with various editing operations in just 4 steps.}
\label{fig:teaser}

\end{strip}

\begin{abstract}
We propose a fast text-guided image editing method called InstantEdit based on the RectifiedFlow framework, which is structured as a few-step editing process that preserves critical content while following closely to textual instructions. Our approach leverages the straight sampling trajectories of RectifiedFlow by
introducing a specialized inversion strategy called PerRFI. 
To maintain consistent while editable results for RectifiedFlow model,
we further propose a novel regeneration method, Inversion Latent Injection, which effectively reuses latent information obtained during inversion to facilitate more coherent and detailed regeneration. Additionally, we propose a Disentangled Prompt Guidance technique to balance editability with detail preservation, and integrate a Canny-conditioned ControlNet to incorporate structural cues and suppress artifacts. Evaluation on the PIE image editing dataset demonstrates that InstantEdit is not only fast but also achieves better qualitative and quantitative results compared to state-of-the-art few-step editing methods. Our code is available here: \url{https://github.com/Supercomputing-System-AI-Lab/InstantEdit}

\end{abstract}
    
\section{Introduction}
\label{sec:intro}

Text-guided image editing has emerged as a powerful tool for creative expression, with diffusion models leading the way in generating high-quality results. The general approach to text-guided image editing includes two steps, \emph{image inversion} and \emph{regeneration}. The inversion process maps the input image into the model's latent space through the reversed generation trajectory~\cite{Song2021DDIM,Ho2020DDPM,ledits++,pnp,null_text_inversion}. The conditional regeneration starts from the inverted noise latent coupled with the editing prompt, allowing modification to happen in the noise space, which then gets mapped back to the image space through the regeneration process. 


Despite showing promising results, the computational demands of text-guided image editing is quite high, creating a significant barrier to real-time interactive editing applications, whereas users expect instant feedback when editing images. Reducing the lengthy denoising steps helps improve the generation speed but generally compromises the generation quality, which motivates a design space of fast and accurate editing techniques to make text-guided image editing more practical for real-world deployment.

There has been many studies of existing literature in reducing the computation cost of the backbone diffusion models, primarily through various distillation techniques to train so called \emph{few-step} diffusion models, which significantly decrease the generation process to less than ten sample steps (e.g., 1-8 steps)~\cite{ldm,sdxl-turbo,instaflow,perflow,lcm}. However, applying these few-step diffusion models for text-guided image-editing raises new challenges:

\begin{itemize}
    \item \textbf{Inaccurate inversion trajectory with few steps.}~Early efforts in text-guided image editing rely on refined DDIM inversion~\cite{null_text_inversion,negative_prompt,pnp}, which is a deterministic sampling process that maps a generated image to a series of noise vectors. However, DDIM inversion becomes accurate only when using a large number of diffusion steps (e.g., more than 20 steps)~\cite{edit-friendly-ddpm}, which is hard to apply to few-step diffusion models.

    \item \textbf{Insufficient editability.} 
The traditional DDIM inversion and its corresponding regeneration method lacks editability in few-step setting. Recent work introduces ``edit-friendly'' DDPM-noise inversion~\cite{edit-friendly-ddpm}, which enables text-based editing with few diffusion steps~\cite{TurboEdit_telaviv}. However, we find that DDPM-noise inversion is still hard to achieve desirable control over edited regions, generating images that loosely follow the target prompt.
    
\end{itemize}

\noindent
Faced with these challenges in few-step text-guided image-editing, we propose \name, a method that enables real-image editing with as few as 8 number of function evaluations (NFE), while achieving high  quality without any fine-tuning.  Instead of relying on the traditional formulation of diffusion model, our key insight is that a sampling process with a linear trajectory incurs a relatively small error in the inversion process. Drawing idea from recent
RectifiedFlow models~\cite{rectifiedflow,perflow,instaflow} that straightens the trajectories from noises to images, we introduces \emph{Piecewise Rectified Flow Inversion (PerRFI)}, which offers descent inversion results while also maintains editability under the few-step editing setup. 

To further mitigate the effect of inaccurate inversion and guarantee sufficient editability for our RectifiedFlow model backbone, \name introduces novel regeneration methods called \emph{Inversion Latent Injection (ILI)} and \emph{Disentangled Prompt Guidance (DPG)}, respectively, which 1) leverage our inverted latent to aid the regeneration process; 2) explicitly computes and scales the orthogonal components between target-prompted and source-prompted conditions,
providing a more accurate guidance signal.
In addition, DPG optionally employs an attention-guided masking mechanism to automatically identify and emphasize regions relevant to the target edit,
and leverages \emph{Canny-conditioned ControlNet}~\cite{zhang2023controlnet} to provide explicit structural guidance throughout both inversion and regeneration. 
This approach ensures faithful preservation of the global layout and local details while allowing for targeted modifications. Importantly, this structural guidance adds minimal computational overhead while  improving the visual quality and consistency of edited results in a flexible manner. 


Through judicious design, \name enables fast and accurate few-step text-guided image editing, with several visual examples shown in \fref{fig:teaser}. Our experiments on the PIE image editing dataset demonstrate the effectiveness and efficiency of \name in improving the image editability. Notably, our approach successfully performs editing in just 8 NFE. As a result, \name not only surpasses other few-step editing frameworks, but also achieves comparable or even better than other multi-step methods in both editability and consistency.

\label{sec:intro}

\begin{figure*}[h!]
 \centering
 \begin{minipage}{\textwidth}
\includegraphics[width=\linewidth]{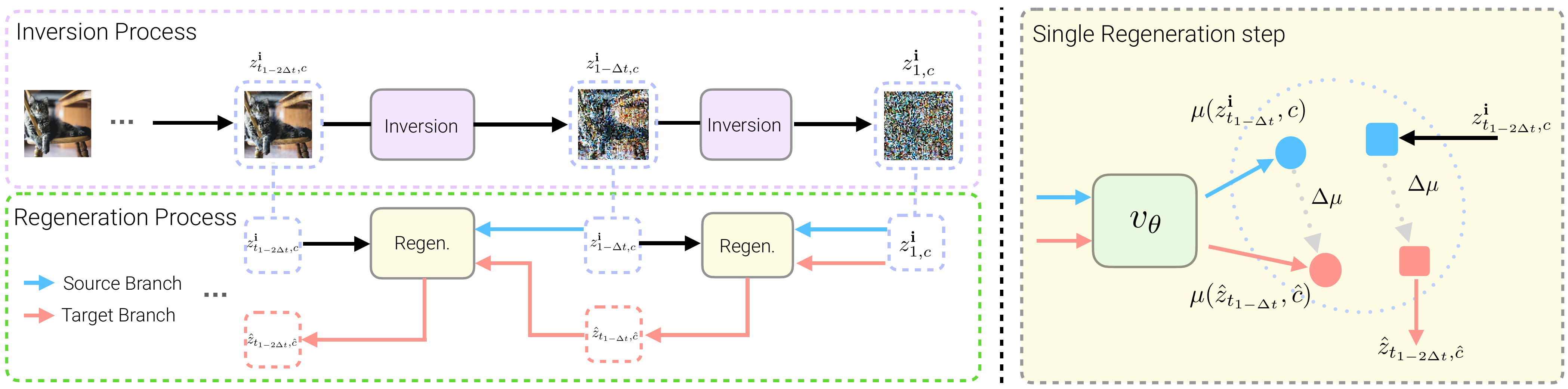}
\caption{{\bf Left}: illustration of the inversion and regeneration process. {\bf Right}: illustration of a single regeneration step. We first calculate the difference between source and target branch output from PerFlow model, and composite this difference back into our stored intermediate latent to generate the final output for this step. 
\label{fig:main fig}
}
\end{minipage}
 \vspace{-15pt}
\end{figure*}
\section{Related Work}
\label{sec:related}

\paragraph{Text-guided image editing.} Text-guided image editing allows users to modify an image based on text descriptions. Early approaches leverage Generative Adversarial Networks (GANs)~\cite{Li2020ManiGAN, Li2020TAGAN, styleclip}. 
While pioneering, those methods often suffer from mode collapse and generate a limited variety of images, and GANs are often unstable and difficult to optimize. Recently, diffusion models represent a newer paradigm in text-guided image editing. Methods such as Prompt2Prompt~\cite{p2p}, MasaCtrl \cite{masactrl}, and Instruct-Pix2Pix \cite{instruct-pix2pix}, have demonstrated improved precision and quality of edits. However, diffusion models also pose challenges, such as the long latency and high generation cost. Different from those efforts, we focus on accelerating the image editing process of diffusion models with high edit quality. 

\paragraph{Few-step image generation.} Few-step image generation aims to reduce the computational cost of multi-step diffusion models. Recent work have tried to tackle this issue by employing a distillation process to fine-tune a multi-step diffusion model to achieve a few-step (e.g.,1--8) schedule. Various distillation objectives have been proposed, such as the adversarial loss~\cite{sdxl-turbo}, distribution matching~\cite{distributed-matching}, consistency objective~\cite{lcm}, and rectified flow~\cite{instaflow,perflow}.
Unlike these existing efforts, our study investigate the interplay between few-step diffusion models and image editing. 

\paragraph{Image editing acceleration.} Researchers have also looked into image editing acceleration. ReNoise~\cite{renoise} proposes to enhance the inversion process by iteratively applying a renoising mechanism at each inversion step, which improves the approximation accuracy of the predicted point along the forward diffusion trajectory. However, the iterative renoising adds additional overhead, which limits its speedups in practice. InfEdit~\cite{inf_edit} introduces an inversion-free method, which eliminates the inversion process through a Denoising Diffusion Consistent Model. TurboEdit~\cite{TurboEdit_telaviv} proposes to adjust the noise schedule for the DDPM-noise inversion and adopts a pseudo-guidance approach to improve the editing strength. Different from previous work, we realize the potential of the RectifiedFlow model with straight sampling trajectories. By designing specific inversion and regeneration strategies for this backbone, we achieve better balance between consistency and editability while still maintaining low generation overhead.

\section{Preliminaries}

\subsection{Diffusion Models}
Diffusion models have a forward and reverse process. The forward diffusion process gradually adds Gaussian noise onto a clean image 
\begin{equation}
    z_t = \sqrt{\alpha_t} z_0 + \sqrt{1 - \alpha_t} \, \epsilon, \quad \epsilon \sim \mathcal{N}(0, I)
\end{equation}
where $z_0$ is from the image distribution, $\alpha_{1:T}$ denotes a variance schedule for time steps $t \sim [1,T]$. During the reverse process, the model tries to remove noise from a noisy image to a clean image. Given a condition $c$ (usually text~\cite{ldm}), a network $\hat{\epsilon}_\theta$ is trained to predict the noise residual $\epsilon_t$, given $z_t$, $c$ and time step $t$ using the objective:
\begin{equation}
    L(\hat{\epsilon}_\theta) = \mathbb{E}_{\epsilon_t \sim \mathcal{N}(0,1)} \left[ \| \hat{\epsilon}_\theta(z_t, c, t) - \epsilon_t \|^2 \right]
\end{equation}

\subsection{RectifiedFlow Diffusion Models}

RectifiedFlow~\cite{rectifiedflow} is a type of flow-based diffusion models, aiming to learn a straightened mapping between the noise distribution $z_1 \sim \pi_1$ and the image distribution $z_0 \sim \pi_0$. 
When start sampling from the end of noise distribution, the sampling trajectory is represented as an ordinary differential equation (ODE):
\begin{equation}
    d {z_t} = v_\theta({z_t}, t) \, dt, \quad t\sim [0, 1].
    \label{rectifiedflow_step}
\end{equation}
Here, rather than predicting noise residual, $v_\theta$ is a velocity field estimator, optimized through a flow matching loss to enforce a linear denoising trajectory.
To numerically solve this ODE, we can discretize the time interval into $N$ small steps with a time increment $\Delta t = \frac{1}{N}$. Let $t_k = k\Delta t$ for $k=\{0, 1, \dots, N\}$, we can iteratively update $z_t$ using:
\begin{equation}
\label{eq:perflow_denoising}
    z_{t_k} = z_{t_{k+1}} + v_\theta(z_{t_{k+1}}, t_{k+1})\Delta t.
\end{equation}



\section{Method}

\subsection{Problem Formulation}

The input to our model is a real image $x$, a source text prompt condition $c$, and a target text condition $\hat{c}$. Users can specify it or derive it from existing image captioning models (\eg, BLIP~\cite{li2023blip}) applied on the input image. The output is an edited image that reflects changes indicated by $\hat{c}$. 
Text-conditional image editing using diffusion models usually consists of two key steps: image inversion and regeneration.

\subsection{Image Inversion}

\begin{figure}[ht!]
 \centering
 \begin{minipage}{\columnwidth}
\includegraphics[width=0.9\linewidth]{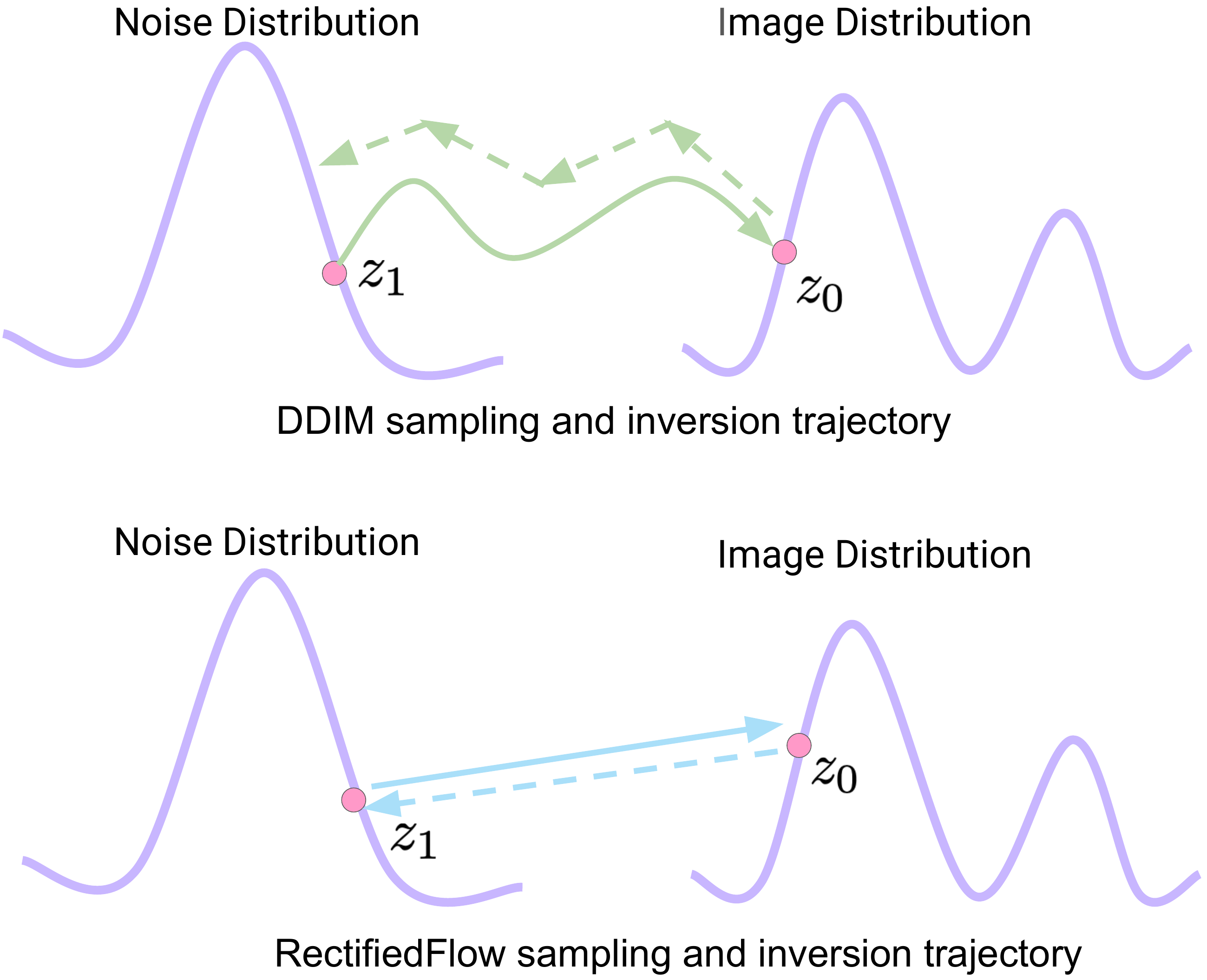}
\caption{Demonstration of the different generation and inversion processes between RectifiedFlow and DDIM-based models. RectifiedFlow enforces a straight sampling trajectory between noise and image space. Its inversion step is fast and incurs less error.
\label{fig:inversion_trajectory}
}
\end{minipage}
 \vspace{-17pt}
\end{figure}

Image inversion sometimes takes the form of image encoding by training an additional encoder to map the input image to editable latents, as operated in~\cite{TurboEdit_adobe}. Instead of building an additional encoder, DDIM inversion starts from $z_0$ and consecutively performs the following to compute inverted latent of the next step:
\begin{equation}
    z_t = \sqrt{\alpha_t} z_0 + \sqrt{1 - \alpha_t} \epsilon_\theta(z_t, t).
\end{equation}
We can find a recursive reference to $z_t$ on the right-hand side, and a common practice is to assume that $\epsilon_\theta(z_t, t) \approx \epsilon_\theta(z_{t-1}, t)$~\cite{Dhariwal_nips_2021}. However, this assumption comes with an underlying condition: we need to enlarge the number of total steps to make each sampling step close to linear. This can be impractical if one cares about the running speed of the editing algorithm. Therefore, in the few-step scenario, DDIM-based inversion easily leads to large inversion error, as shown in \fref{fig:inversion_trajectory}. 
Alternatively, DDPM-noise inversion methods ~\cite{TurboEdit_telaviv,edit-friendly-ddpm} iteratively add noise to the latent in the last step to substitute the inversion process. Although simple, this cannot guarantee that the derived latent fall on the optimal editing trajectories, and we experimentally find that this approach shows limited editability strength.

\begin{figure}[h!]
 \centering
 \begin{minipage}{\columnwidth}
\includegraphics[width=\linewidth]{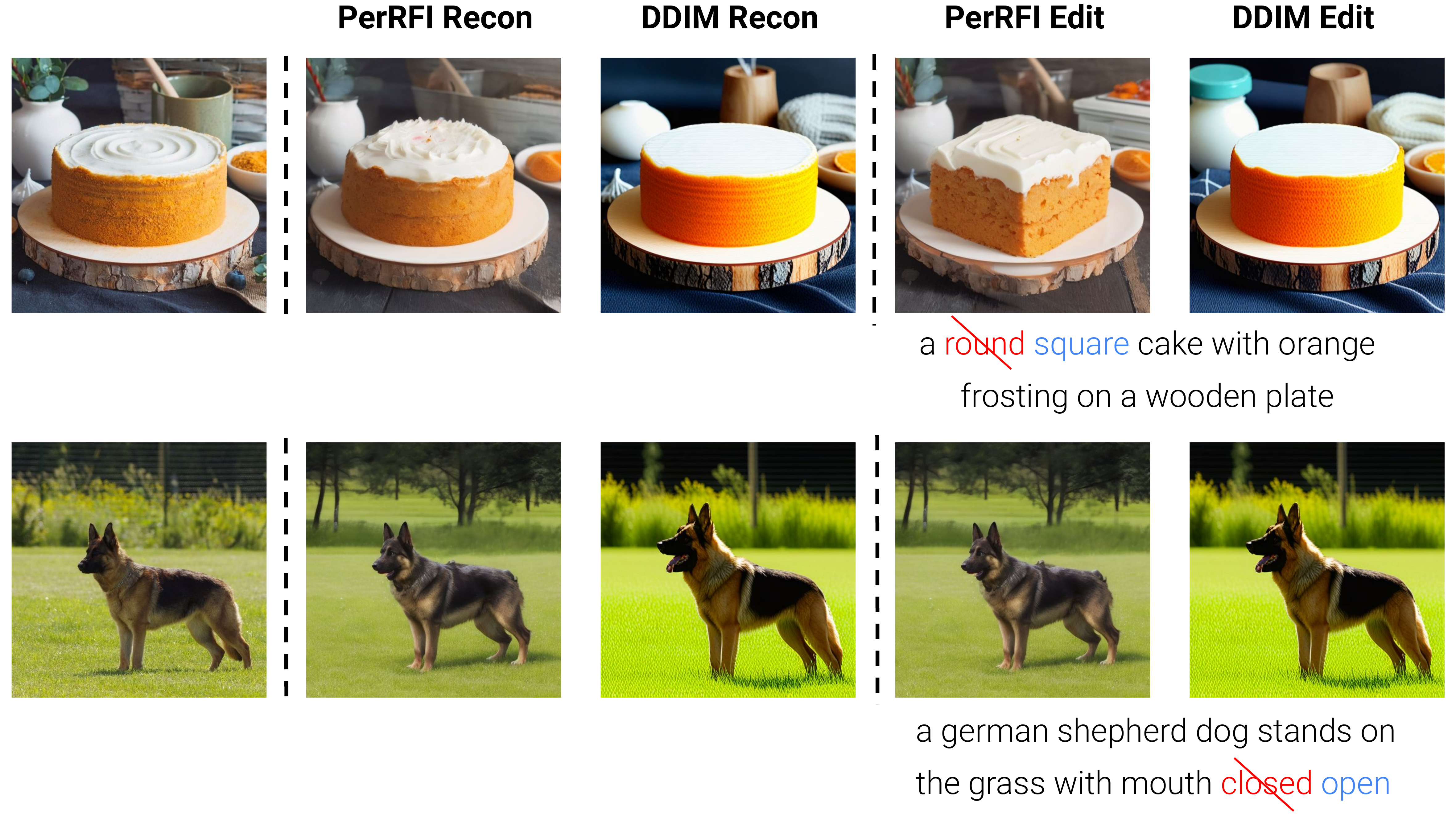}
\caption{Visual results for image reconstruction and editing with vanilla PerRFI and DDIM Inversion, with other techniques intact. The backbone for DDIM inversion is SDXL-Turbo. PerRFI consistently shows better image quality comparing with DDIM inversion. 
\label{fig:inversion_motivation}
}
\end{minipage}
 \vspace{-15pt}
\end{figure}

\paragraph{Piecewise Rectified Flow Inversion (PerRFI).}
Inspired by the recent success of RectifiedFlow based approaches~\cite{rectifiedflow} in straightening the denoising trajectory, we propose to invert images using a recently published RectifiedFlow model---PeRFlow~\cite{perflow}. PeRFlow divides the sampling process into multiple time windows and straightens the trajectories in each window via the ReFlow operation, achieving piece-wise  linear flows. Specifically, borrowing the context of Eq.~\ref{eq:perflow_denoising} which performs the denoising process using the velocity function $v_{\theta}$, the inversion is simply driven by:
\begin{equation}
    \label{eq:perflow_diffusion}
    z_{t_{k+1}, c}^{\mathbf{i}} = z_{t_k,c}^{\mathbf{i}} - v_\theta(z_{t_k,c}^{\mathbf{i}}, t_k, c)\Delta t,
\end{equation} where $z_{t_k,c}^{\mathbf{i}}$ indicates the inverted latent of time step $t_k$ under condition $c$. Given that PeRFlow's noise trajectories are straight lines, this inversion has relatively small approximation errors. Therefore, it demonstrates better image quality in reconstructions and editing, as in Fig.~\ref{fig:inversion_motivation}, compared to DDIM inversion. Note that our inversion approach is not affixed to PeRFlow, and can be easily applied to other RectifiedFlow methods.

\subsection{Regeneration}
PerRFI alone does not create the most satisfying results. To further mitigate the effect of inversion error, while achieving better editability, we innovate our regeneration pipeline in two directions: the sampling strategy and the guidance method, which we name as \textbf{Inversion Latent Injection (ILI)} and \textbf{Disentangled Prompt Guidance (DPG)}.
\paragraph{Inversion Latent Injection.}~The simplest regeneration can be executed through the denoising process described in Eq.~\ref{eq:perflow_denoising} from an initial inverted latent $z_{t_k,c}^{\mathbf{i}}$ ($t_k$ doesn't have to be 1) all the way to the clean image space, but swap the text condition from $c$ to $\hat{c}$:
\begin{equation}
    \hat{z}_{t_k, \hat{c}} = \hat{z}_{t_{k+1}, \hat{c}} + v_\theta(\hat{z}_{t_{k+1}, \hat{c}}, t_{k+1}, \hat{c})\Delta t.
\end{equation}
This process does not utilize any intermediate inverted latent other than $z_{t_k,c}^{\mathbf{i}}$; thus, we name it No Latent Injection (NLI). The final generated results from this method can differ significantly from the input image, as errors may accumulate in $z_{t_k,c}^{\mathbf{i}}$ throughout the inversion steps, causing deviation from the ideal trajectory.

On the other hand, DDPM-noise inversion~\cite{edit-friendly-ddpm,TurboEdit_telaviv} injects scheduled DDPM noise into the latent image and uses it as the unconditional intermediate latent; hence, we call this method Noise Latent Injection (NSLI). However, the scheduled, nondeterministic DDPM noise causes the image latent to diverge from its normal ODE trajectory, introducing incoherent modifications that make it difficult to align precisely with the target prompt.
To tackle the above issues, we propose our regeneration pipeline, Inversion Latent Injection (ILI). When doing inversion, we will store all the intermediate inverted latent from PerRFI and reuse them to calibrate each regeneration step:
\begin{equation}
\label{eq:ef++}
   \hat{z}_{t_k, \hat{c}} =z_{t_{k}, c}^{\mathbf{i}} + \mu(\hat{z}_{t_{k+1}},\hat{c})-\mu(z^{\mathbf{i}}_{t_{k+1}},c)
\end{equation}
where $\mu(z, c)=z + v_{\theta}(z, c)$, a one step denoised latent of $z$ given condition $c$. The intuition is that inverted latent at earlier time steps (close to clean image) accumulate less error during inversion, i.e. $z_{t_{k-1},c}^{\mathbf{i}}$  is more accurate than $z_{t_k,c}^{\mathbf{i}}$. Every time we calculate one step of denoising, we anchor back to the stored latent to prevent error accumulation. 

\paragraph{Disentangled Prompt Guidance.}~Note that the later part of Eq.~\ref{eq:ef++} can be further expanded into:
{\small
\begin{equation}
\label{eq:turboedit}
\begin{aligned}
\mu(\hat{z}_{t_{k+1}},\hat{c})-\mu(z^{\mathbf{i}}_{t_{k+1}},c) &= 
\underbrace{\mu(\hat{z}_{t_{k+1}},\hat{c})-\mu(\hat{z}_{t_{k+1}},c)}_{\text{cross-prompt}} \\
&+ \underbrace{\mu(\hat{z}_{t_{k+1}},c) - \mu(z^{\mathbf{i}}_{t_{k+1}},c)}_{\text{cross-trajectory}},
\end{aligned}
\end{equation}
}where the first cross-prompt term captures the difference between predictions for the generation trajectory under the new and original prompts. The second term is the difference between predictions from the new trajectory and the original one under the same prompt. 
TurboEdit \cite{TurboEdit_telaviv} finds that scaling the cross-prompt term provides a good guidance towards the target prompt, which they termed as Pseudo-Guidance (PG).


After experimenting on TurboEdit, we observe that scaling the cross-prompt term can induce undesirable changes in the generated image (see Fig.~\ref{fig:guidance_visual}). We hypothesize that this issue arises primarily from the use of Pseudo-Guidance. Notably, the term $\mu(\hat{z}_{t_{k+1}},c)$ in the cross-prompt formulation is influenced by $\hat{z}_{t_{k+1}}$, which is a latent state strongly conditioned on the target prompt. As a result, it becomes challenging for the source prompt to provide accurate guidance in the latent space dominated by the target prompt’s influence. To address this, we propose increasing the disentanglement between the guidance signals of the target and source prompts, mitigating the impact of inaccurate guidance from the source prompt. We first reformulate the Pseudo-Guidance under our generation setting as 
\begin{equation}
    \text{PG}: w(v_\theta(\hat{z}_{t_{k+1}},\hat{c})-v_\theta(\hat{z}_{t_{k+1}},c))
\end{equation}
where $w$ is the scaling factor. In order to obtain better disentanglement, we scale the component of the target signal that is orthogonal to the source signal, which we term the method as Disentangled Pormpt Guidance (DPG).
\begin{equation}
    \text{DPG}: w(v_\theta(\hat{z}_{t_{k+1}},\hat{c})-\text{Proj}(\hat{c},c))
\end{equation}
where $\text{Proj}(\hat{c},c)$ is 
defined as 
\begin{equation}
    \text{Proj}(\hat{c},c) = \frac{v_\theta(\hat{z}_{t_{k+1}},\hat{c}) \boldsymbol{\cdot} v_\theta(\hat{z}_{t_{k+1}},c)}{||v_\theta(\hat{z}_{t_{k+1}},c)||}.
\end{equation}
Here, $\boldsymbol{\cdot}$ is the dot product operation; $||\dots||$ represents the norm of the vector; $w$ is the scaling factor. Intuitively, this approach enables the regeneration process to filter out some disturbance from the inaccurate guidance by source prompt, improving background preservation while maintaining high editability from the PG schedule.

We can optionally leverage an \textbf{Attention Masking mechanism} based on the original and target prompts to further disentangle the effect of target and source prompt. We can identify the editing word $w_e$ by finding the difference between the source prompt and the target prompt. At time $t_k$, we obtain the mask as:
\begin{equation}
    m = \text{Threshold}[A_{t_k}(w_e),\alpha]
    \label{mask}
\end{equation}
$m$ is the binary mask obtained from the averaged cross attention map $ A_{t_k}(w_e)$ , following the setting in ~\cite{masactrl},
with threshold parameter $\alpha$. The part with value above threshold will be assigned as 1, otherwise 0, so we mask out the region we do not want to perform editing. The final formulation for DPG with mask is
\begin{equation}
\scalebox{1.0}{
$ 
\text{DPG}: m \odot w(v_\theta(\hat{z}_{t_{k+1}},\hat{c})-\text{Proj}(\hat{c},c))
$
}
\end{equation}
which means that we only perform guidance within the masked region. For better disentanglement, we also apply the same mask to the cross-trajectory part.

\subsection{ControlNet Guided Editing}
To further preserve background and minimize structural information loss, we develop a plug-and-play method which replaces the backbone network with Canny-conditioned ControlNet~\cite{zhang2023controlnet}. Canny edges can be extracted quickly with only marginal computation overhead. With edge information inserted, we find improvements in performing more accurate image inversion and thus reducing structural information loss. Another advantage of this method is that users can easily control the structural rigidity by adjusting the ControlNet conditioning scale, which is supported by most of the existing ControlNet pipelines. 

\begin{figure*}[h!]
 \centering
 \begin{minipage}{\textwidth}
\includegraphics[width=\linewidth]{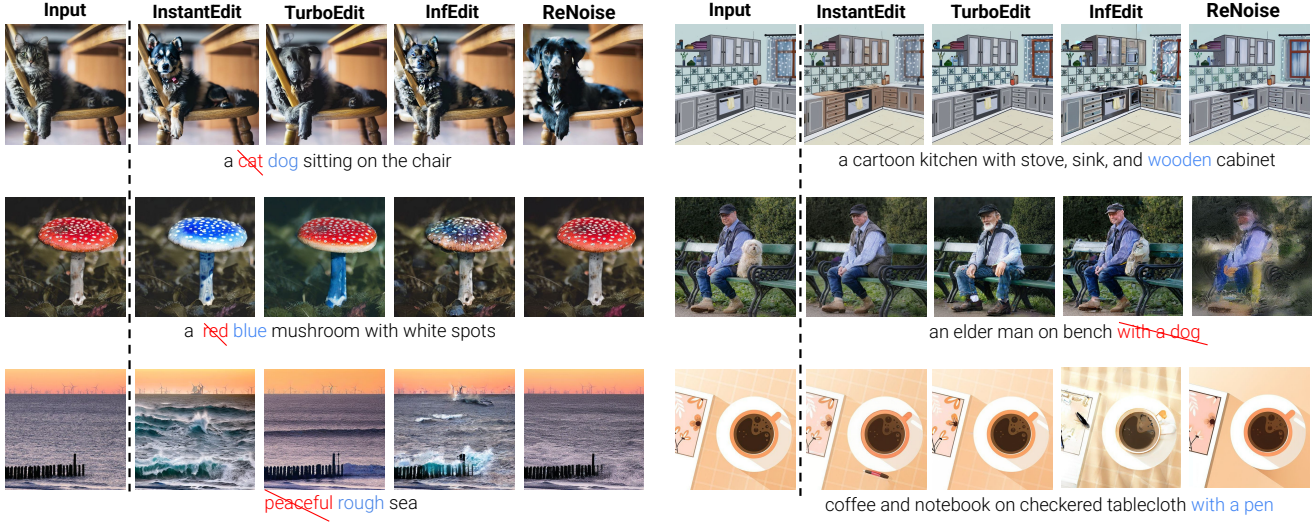}
\caption{ Qualitative comparison between \name and other few-step editing baselines. InstantEdit shows good alignment with the editing prompt while also maintaining consistency with the original image in varies editing settings.
\label{fig:main_fig}
}
\end{minipage}
\end{figure*}
\begin{table*}[h!]
    \centering
    \setlength{\tabcolsep}{3pt}
    \renewcommand{\arraystretch}{1.1}
    \begin{tabular}{l|c|cccc|cc|ccc}
        \toprule
        \textbf{Method} & \textbf{Structure}& \multicolumn{4}{c}{\textbf{Consistency}} & \multicolumn{2}{c}{\textbf{Alignment}} & \multicolumn{3}{c}{\textbf{Efficiency}} \\
        \midrule
            & 
        \textbf{Distance}$_{10^3}^ \downarrow$ & 
        \textbf{PSNR}$\uparrow$ & \textbf{LPIPS}$_{10^3}^\downarrow$ & \textbf{MSE}$_{10^4}^\downarrow$ & \textbf{SSIM}$_{10^2}^\uparrow$ &
        \textbf{Whole}$\uparrow$ & \textbf{Edited}$\uparrow$ &
        \textbf{Time}$\downarrow$ & \textbf{NFE}$\downarrow$ & \textbf{Step}$\downarrow$\\
        \midrule
        \textbf{EF} & \textbf{8.39} & 27.49 & 44.38 & 29.79 & 85.61 & 25.87 & 22.14 & 32.46 & 70 & 50\\
         \textbf{ProxG} & \textbf{8.39}& 28.45 & 38.27 & 25.63 & 85.87 & 25.04 & 21.64 & 17.95 & 100 & 50\\ 
         \textbf{P2P} & 9.58& 27.72 & 44.98 & 30.02 & 85.01 & 24.94 & 21.57 & 92.14 & 100 & 50\\ 
        \textbf{DI} & 11.60 & 27.25 & 49.25 & 32.87 & 84.86 & 25.83 & 22.39 & 30.90 & 100 & 50\\
        \textbf{InfEdit} & 13.87 & 28.63 & 39.80 & 33.19 & 86.28 & 25.84 & 22.44 & \textbf{2.12} & 12 & 12\\
        \textbf{\name (Ours)} & 12.57 & \textbf{29.63} & \textbf{35.27} & \textbf{24.57} & \textbf{87.40} & \textbf{26.06} & \textbf{22.73} & 3.67 & 24 & 12\\

        \midrule
        \textbf{ReNoise} & 20.31 & 24.89 & 83.26 & 52.9 & 82.12 & 25.26 & 21.68 & 2.47 & 16 & 4\\
        \textbf{TurboEdit} & 18.57 & 24.59 & 77.53 & 58.48 & 82.64 & 25.7 & 22.3 & 0.96 & 4 & 4\\
        \textbf{InfEdit} & {\bf 16.19} & 26.75 & 50.79 & 42.33 & 84.71 & 25.68 & 22.27 & \textbf{0.80} & 4 & 4\\
        \textbf{\name (Ours)} & 17.14 & {\bf 27.96} & \textbf{44.39} & \textbf{34.94} & \textbf{86.44} & \textbf{26.28} & \textbf{22.82} & 1.37 & 8 & 4 \\ \hline
    \end{tabular}
    \caption{Comparison of different image editing method  on PIE Bench. The top group shows methods that run in multi-step editing case. The bottom group shows method that run in few-step scenario. TurboEdit is specfically designed for SDXL-Turbo; EF is based on SD 1.4; all other models are based on/distilled from SD 1.5.}
    \label{tab:comparison}
\end{table*}
\section{Experiments}
\label{sec:eval}

\subsection{Evaluation Methodology}

\paragraph{Implementation} We implement \name based on the model pipelines built in Diffusers~\cite{diffusers}. In this work, we use PeRFlow as the backbone that is distilled from Stable Diffusion 1.5 (SD1.5) ~\cite{sd1.5}. One important note is that there exists a tradeoff between the consistency metrics (Structure, Consistency) and the editability metrics (Alignment). In our method, the important parameters that control this tradeoff are the ControlNet conditioning scale and the DPG scale. We refer readers to the supplementary material for the detailed hyperparameters and the selection process for these parameters.

\paragraph{Benchmarks} We use an established benchmark, PIE Bench~\cite{pnp}, which focuses on 9 editing types: \emph{change object, add object, delete object, change content, change pose, change color, change material, change background}, and \emph{change style}.

\paragraph{Evaluation Metrics} We follow the setting from Ju et al.~\cite{pnp} to evaluate on:
\begin{itemize}
    \item \textbf{Structure Preservation.} We use the structure distance \cite{structure_distance} to capture the magnitude of structural change while ignoring appearance information.
    \item \textbf{Consistency.} We calculate Mean Squared Error (MSE), Peak Signal-to-Noise Ratio (PSNR), Structural Similarity Index Measure (SSIM) \cite{ssim}, and Learned Perceptual Image Patch Similarity (LPIPS)~\cite{lpips} on the part excluded by the editing mask to evaluate the overall consistency of the unedited area.
    \item \textbf{Image-Prompt Alignment.} We use CLIPScore~\cite{clipscore} to calculate the similarity between target prompt and 1) the whole image; 2) only the edited part indicated by mask. This metric reflects the editability of the model.
    \item \textbf{Efficiency.} We calculate the wall clock time for each method to process a single image. We also include the number of function evaluation (NFE), which indicates the total number of forward passes into the model when editing a single image. Another metric in this category is Step, which only refers to the sampling length of the model. We include Step since it is also a common expression used in some previous work ~\cite{p2p,masactrl}.   
\end{itemize} 

\subsection{Main Results}

In this part, we compare \name with the following few-step editing baselines: 1) ReNoise~\cite{renoise}; 2) InfEdit~\cite{inf_edit}; 3) TurboEdit~\cite{TurboEdit_telaviv}. We also include multi-step editing methods: 1) Edit-Friendly DDPM Inversion(EF)~\cite{edit-friendly-ddpm} 2) Proximal Guidance(ProxG)~\cite{prox_guide} 3) Prompt-to-Prompt+Null-text Inversion(P2P)~\cite{p2p,null_text_inversion} 4) Direct Inversion(DI)~\cite{pnp}. We further test InfEdit in the default setting of 12 steps, and to compare with it, we also run \name in 12 steps to demonstrate the performance of our approach in the multi-step editing scenario. 

\paragraph{Quantitative Results}
As shown in \tref{tab:comparison}, although our method has a little time overhead compared to InfEdit and TurboEdit due to the relatively more time-consuming inversion process, it surpasses all other baselines in almost all of the metrics in both the few-step and the multi-step cases. We observe that for \name and InfEdit, when increasing the generation steps, consistency and structure scores improve largely with alignment metrics maintaining the same level as the few-step setting.

\paragraph{Qualitative Results}
 We show qualitative comparison of image editing results between \name and other methods in \fref{fig:main_fig}. While all methods show certain editability, \name achieves better alignment with the editing prompts and also obtains better consistency with the original images in edited area. As an example, for the image of a dog, \name offers the best edited result without noticeable information loss of background regions, but TurboEdit and InfEdit fail to show a reasonable dog and ReNoise misses the structure of the chair.

\paragraph{User Study} In this section, we conduct a user study to reflect the overall visual quality of different few-step editing methods. We randomly sampled 15 images from PIE Bench and run on 4 methods: TurboEdit, InfEdit, ReNoise, and InstantEdit. Users will choose the best edited image among the 4 based on: 1) Editability; 2) Consistency; 3) Visual quality. We give a detailed elaboration on how we conduct this survey in the supplementary material. We collect responses from 37 users and result in a total of 545 effective answers. The result shows in \tref{tab:user_study}. In general, \name and TurboEdit are more preferable with our \name being the most frequently selected method. We notice that the user study shows some inconsistency with the quantitative result shown in \tref{tab:comparison}. While InfEdit shows more promising quantitative results than TurboEdit, it is less favored by users. We examine the visual results produced by these two methods and find that InfEdit tends to generate small artifacts and distortions on images which are largely ignored during the metric calculation, but will likely be captured by human users. Please refer to the supplementary materials for images chosen for user study and further analysis.

\begin{table}[ht!]
    \centering
    \setlength{\tabcolsep}{4pt}
    \renewcommand{\arraystretch}{1.1}
    \begin{tabular}{lcc}
        \toprule
        \textbf{Method} & \textbf{No. selection} & \textbf{Percentage (\%)} \\
         \midrule
        \textbf{ReNoise} & 50 &9.17 \\ 
        \textbf{InfEdit} & 100 & 18.35 \\ 
        \textbf{TurboEdit} & 191 & 35.05 \\ \hline
        \textbf{\name} & 204 & 37.43 \\ \hline
    \end{tabular}
    \caption{User study result. We report the number of selections of each method by users from the worst to the best.}
    \label{tab:user_study}
\end{table}

\begin{table*}[h!]
    \centering
    \setlength{\tabcolsep}{3pt}
    \renewcommand{\arraystretch}{1.1}
    \label{tab:main_result}
    \begin{tabular}{lccccccc}
       \toprule
        & 
        \textbf{Distance}$_{10^3}^ \downarrow$ & 
        \textbf{PSNR}$\uparrow$ & \textbf{LPIPS}$_{10^3}^\downarrow$ & \textbf{MSE}$_{10^4}^\downarrow$ & \textbf{SSIM}$_{10^2}^\uparrow$ &
        \textbf{Whole}$\uparrow$ & \textbf{Edited}$\uparrow$ \\
        \midrule
        \textbf{DDIM} & 79.42 & 18.83 & 173.47 & 179.45 & 66.53 & 26.93 & 23.57 \\
        \textbf{PerRFI} & 48.51 & 21.10 & 157.21 & 106.42 & 76.34 & 27.00 & 23.74 \\ \hline
    \end{tabular}
    \caption{Quantitative results for reconstruction with DDIM inversion and PerRFI.}
    \label{tab:inversion}
\end{table*}

\begin{table*}
 \centering
\setlength{\tabcolsep}{3pt}
\renewcommand{\arraystretch}{1.1}
\label{tab:regen_comparison}
\begin{tabular}{lccccccc|c}
   \toprule
    &
    \textbf{Distance}$_{10^3}^ \downarrow$ & 
    \textbf{PSNR}$\uparrow$ & \textbf{LPIPS}$_{10^3}^\downarrow$ & \textbf{MSE}$_{10^4}^\downarrow$ & \textbf{SSIM}$_{10^2}^\uparrow$ &
    \textbf{Whole}$\uparrow$ & \textbf{Edited}$\uparrow$ & \textbf{Components} \\
    \midrule
    \textbf{ILI $\Rightarrow$NSLI} & 20.61 & 27.02 & 47.65 & 46.62 & 85.64 & 25.92 & 22.32 & \multirow{1}{*}{Regeneration} \\
    \midrule
    \textbf{DPG $\Rightarrow$ PG} & 19.86 & 24.19 & 70.82 & 60.00 & 83.92 & 25.85 & 22.15 & \multirow{2}{*}{Guidance} \\
    \textbf{DPG $\Rightarrow$ DPG(w/o mask)} & 16.97 & 24.79 & 65.68 & 51.52 & 84.44 & 25.83 & 22.25 \\
    \midrule
    \textbf{- ControlNet} & 24.31 & 26.68 & 57.18 & 49.23 & 84.98 & 26.31 & 22.79 & \multirow{1}{*}{ControlNet} \\
    \midrule
    \textbf{InstantEdit} & 17.14 & 27.96 & 44.39 & 34.94 & 86.44 & 26.28 & 22.82 & \multirow{1}{*}{Full model}\\
    \bottomrule
    \end{tabular}
    \caption{Quantitative results for the compilation of ablation experiments, including regeneration, guidance, and Canny-conditioned ControlNet}
    \label{tab:ablation_compile}
\end{table*}

\begin{figure}[ht!]
 \centering
 \begin{minipage}{\columnwidth}
\includegraphics[width=\linewidth]{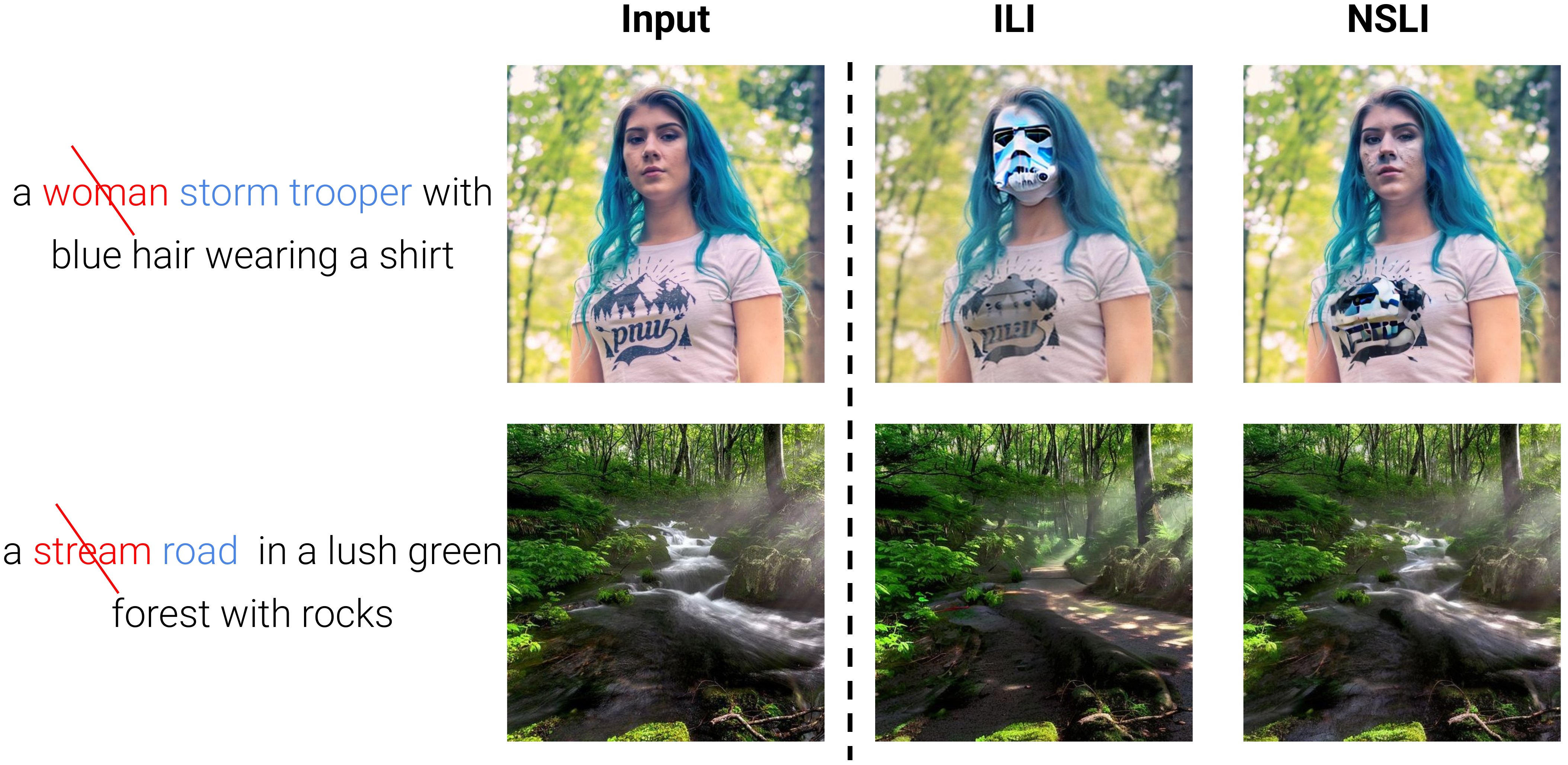}
\caption{Qualitative comparison between ILI and NSLI. ILI leverages intermediate latent closely follow the generation trajectory of the image, which results in better editability. 
\label{fig:inversion_visual}
}
\end{minipage}
 \vspace{-15pt}
\end{figure}

\begin{figure}[ht!]
 \centering
 \begin{minipage}{\columnwidth}
\includegraphics[width=\linewidth]{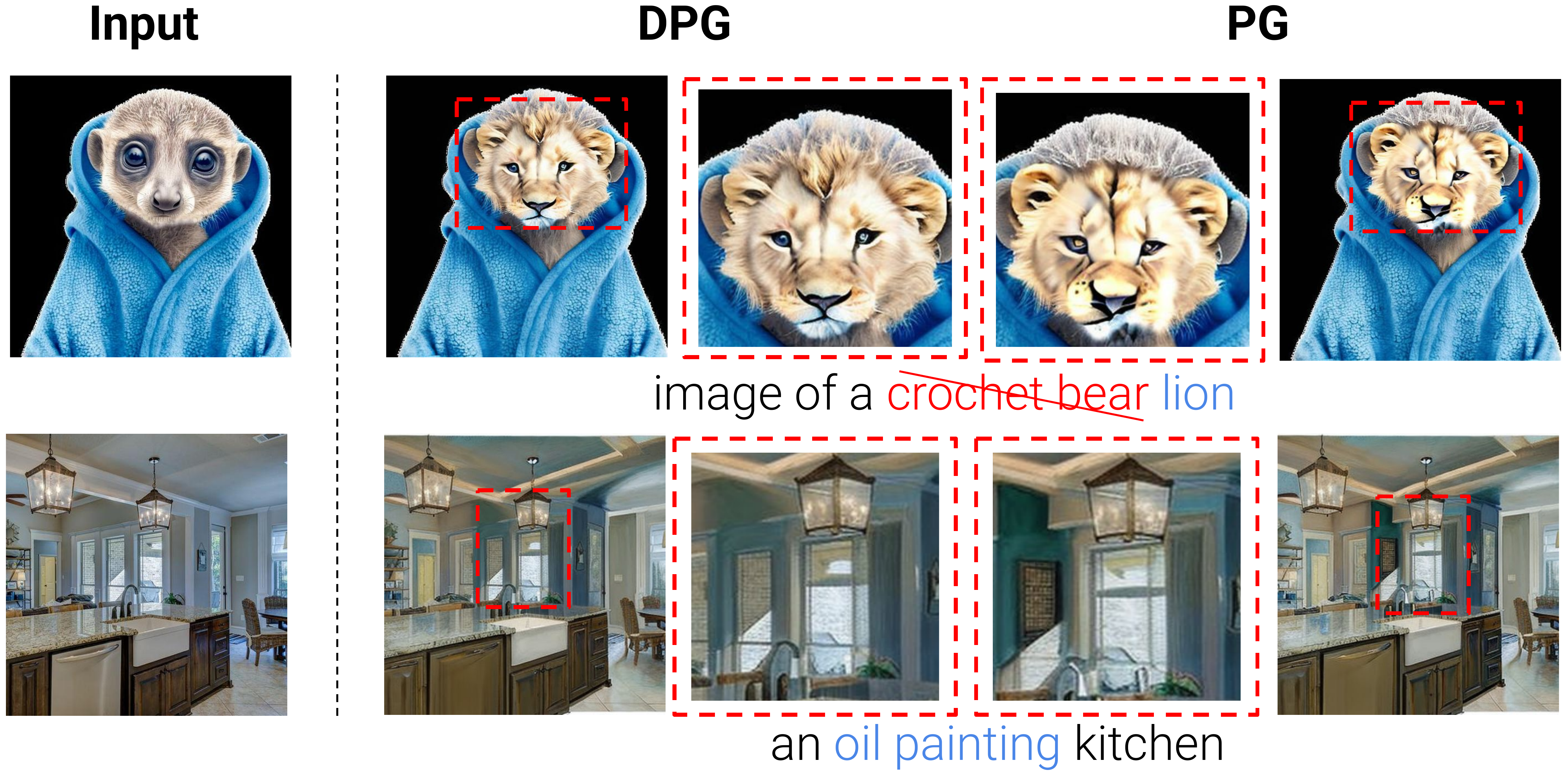}
\caption{ Qualitative comparison between Disentangled Prompt Guidance (DPG) and Pseudo-Guidance (PG). The orthogonal calculation in DPG helps to alleviate unwanted changes including the lion face on row 1 and the window on row 2. 
\label{fig:guidance_visual}
}
\end{minipage}
\end{figure}

\begin{figure}[ht!]
 \centering
 \begin{minipage}{\columnwidth}
\includegraphics[width=\linewidth]{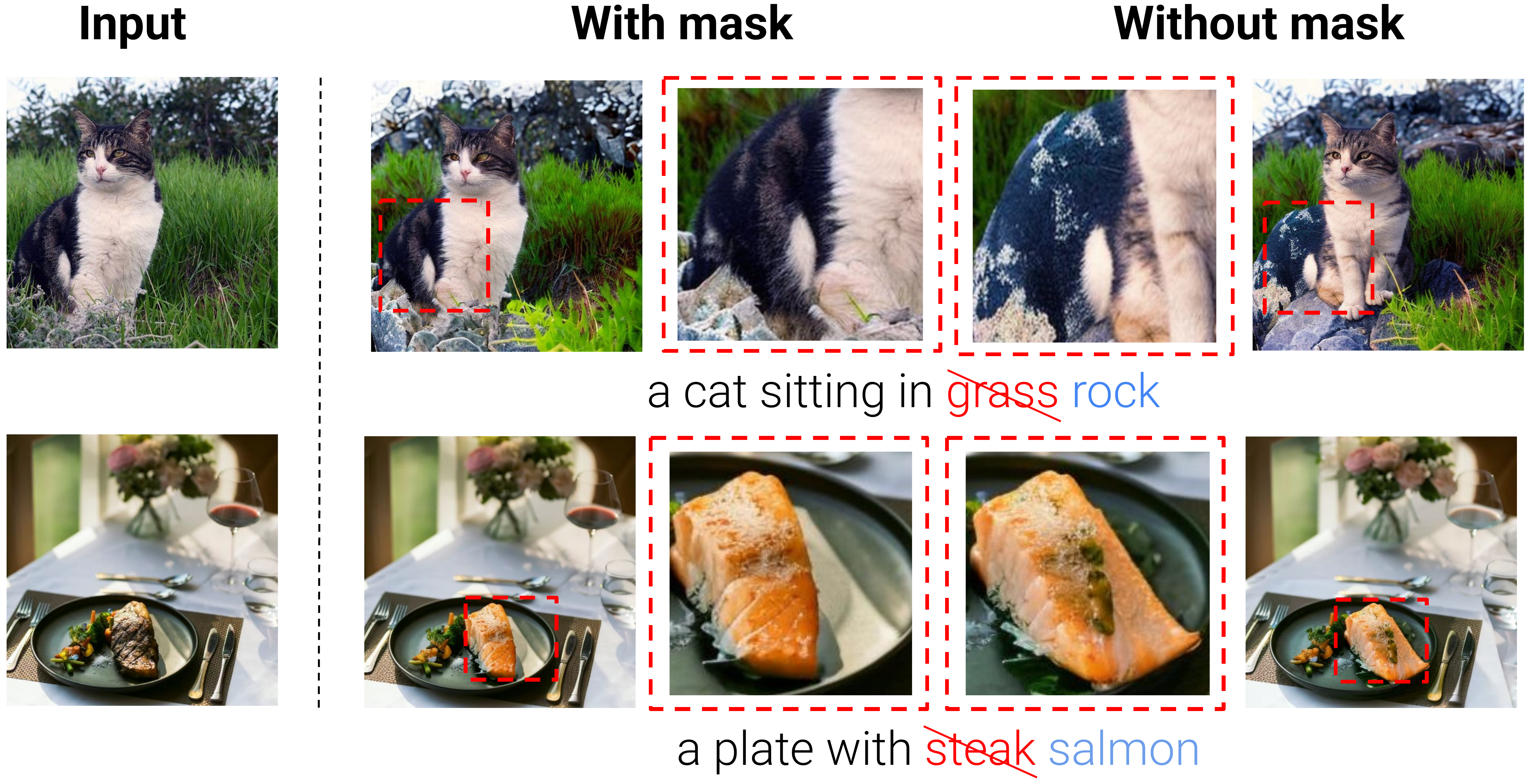}
\caption{Qualitative comparison for DPG with and without attention masking. Attention masking further disentangles the effect of source and target prompts which prevents unwanted editing in unrelated regions.
\label{fig:masK_visual}
}
\end{minipage}
\end{figure}

\begin{figure}
 \centering
 \begin{minipage}{\columnwidth}
\includegraphics[width=\linewidth]{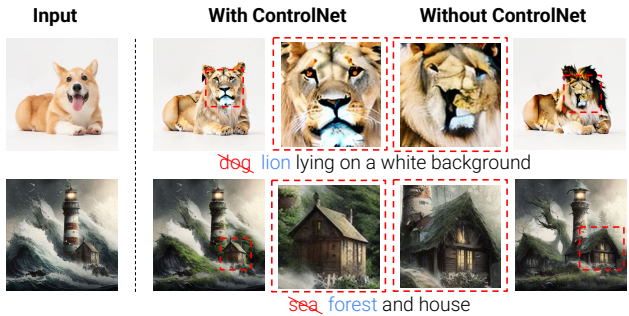}
\caption{ Qualitative comparison for results with and without Canny-conditioned ControlNet.
ControlNet helps prevent structural information loss during inversion and generation process to avoid unwanted structural change and distortions.
\label{fig:controlnet_visual}
}
\end{minipage}
 \vspace{-15pt}
\end{figure}

\subsection{Ablation Studies}

We study how each component in \name contributes to the editing results by 1) performing inter-comparison with the counterpart methods for PerRFI, ILI, and DPG. 2) performing intra-comparision for Canny-conditioned ControlNet. We provide more detailed hyperparameter ablation for ControlNet scale and attention masking threshold in supplementary material.
\begin{itemize}
    \item \textbf{PerRFI vs. DDIM Inversion.}~We compare the image reconstruction performance of PerRFI and DDIM Inversion, where the latter operates on SDXL-Turbo. The quantitative results are presented in \tref{tab:inversion}, while qualitative comparisons are shown in \fref{fig:inversion_motivation}. To ensure a fair comparison, all other techniques remain consistent across both methods. Notably, the CLIPScore in this experiment evaluates the alignment between the generated images and the original prompts, rather than the target prompts used for editing.
    \item \textbf{ILI vs. NSLI.}~We compare our regeneration method ILI with the major counterpart NSLI. NSLI uses the noised image latent from DDPM-noise inversion, while ILI leverages the intermediate inverted latent from our PerRFI. We can seamlessly replace ILI with NSLI in our pipeline to ensure a fair comparison. From the Regeneration section in \tref{tab:ablation_compile} and \fref{fig:inversion_visual}, we show that our regeneration method achieves better results in consistency metrics and, more notably, better prompt-image alignment.
    \item \textbf{DPG vs. PG.}~Pseudo-Guidance(PG) scales the cross-prompt components; while DPG scales the orthogonal components between target and source guidance which provides better disentanglement between the two signals to filter out inaccurate guidance signal from source prompt. DPG can also leverage the attention masking mechanism to provide further disentanglement. To perform this ablation, Pseudo-Guidance can also be seamlessly plugged into our pipeline to replace our guidance method.
    From the Guidance section in \fref{fig:guidance_visual} and the Guidance section in \tref{tab:ablation_compile}, we visually and quantitatively show that our Disentangled-Prompt-Guidance achieves better structural consistency, while maintaining the model editability. In \fref{fig:masK_visual}, we show the qualitative effect of attention masking, and we refer readers to supplementary material for further investigation of this method when we apply it to other baseline models.
    \item \textbf{Canny-Conditioned ControlNet.}~Starting from our final equipment, we remove the plug-and-play Canny-conditioned ControlNet, and show how this component affects our performance under the consistency-editability tradeoff. The result is depicted in the ControlNet section in  \tref{tab:ablation_compile}. By adding ControlNet, we enventually achieve a better balance between consistency and editability. We also provide visual results to demonstrate the effect of ControlNet in Fig ~\ref{fig:controlnet_visual}. The figure reveals that ControlNet helps prevent structural information loss during inversion and generation process to avoid unwanted structural changes and distortions.
\end{itemize}


\section{Conclusion}

In this work, we present \name, a fast and accurate text-guided image editing method that leverages RectifiedFlow model to improve the inversion accuracy in few-step diffusion process while proposing novel techniques including Inversion Latent Injection and Disentangled Prompt Guidance to enhance image consistency and model editability. We further use Canny-conditioned ControlNet to better preserve structure information of edited images.Together, \name achieves higher image editing quality than alternative methods while still enjoying fast editing speed. However, \name still faces several limitations. Firstly, due to the inversion method, \name still creates small time overhead compared to InfEdit and TurboEdit. Next, \name is still constrained to moderate edits and may encounter challenges in the case of large structural change, such as pose modification. However, this setting is in general a very difficult task with only textual guidance. Existing works like MasaCtrl ~\cite{masactrl} and InfEdit ~\cite{inf_edit} requires complicated attention manipulation and multi-step editing to achieve slight structure modification. Another line of work requires additional guidance signals, including drag points and regions \cite{dragdiffusion, dragondiffusion, geodiffuser,dragnoise}. We plan to take these as our future work to perform more flexible and faster text-guided image editing. 


\section*{Acknowledgments}

We sincerely appreciate the insightful feedback from the anonymous reviewers. This research was supported by the National Science Foundation (NSF) under Grant No. 2441601. The work utilized the Delta and DeltaAI system at the National Center for Supercomputing Applications (NCSA) and Jetstream2 at Indiana University through allocation CIS240055 from the Advanced Cyberinfrastructure Coordination Ecosystem: Services \& Support (ACCESS) program, which is supported by National Science Foundation grants \#2138259, \#2138286, \#2138307, \#2137603, and \#2138296. The Delta advanced computing resource is a collaborative effort between the University of Illinois Urbana-Champaign and NCSA, supported by the NSF (award OAC 2005572) and the State of Illinois. UIUC SSAIL Lab is supported by research funding and gift from Google, IBM, and AMD. 
{
    \small
    \bibliographystyle{ieeenat_fullname}
    \bibliography{reference}
}

\clearpage
\setcounter{page}{1}
\appendix
\section{Hyperparameters and Design Choice}
Here we present our hyperparamater selection for the main results for reproducibility. For ControlNet, the ControlNet conditioning scale is set to 0.4. In the inversion process, we do not use classifier-free guidance (CFG); while in regeneration, there are two vital parameters, the DPG guidance scale, which is set to 2.5, and the attention mask threshold, which is 0.4. For the attention mask implementation, we aggregate the cross attention maps with dimension $16 \times 16$ and extrapolate to the dimension of the latent, then we perform the thresholding operation.

\section{Consistency-Editability Tradeoff}
Consistency-editability tradeoff is a commonly recognized property in the setting of image editing as discussed by previous work such as ReNoise and InfEdit and we demonstrate it in \tref{tab:supp_instant_dpg_scale},~\tref{tab:supp_infedit_cfg_scale},~\tref{tab:supp_turbo_pg_scale},~\tref{tab:supp_renoise_cfg_scale}. This also affects how we choose our hyperparameters for comparisons. For example, one method can adjust the hyperparameter (e.g. guidance scale) to achieve higher editability above other methods with a sacrifice on its consistency metrics, which makes the result hard to interpret. Therefore, we propose to adjust the hyperparameters to align on one type of metric and compare the overall performance across all other metrics. In Tabel 1 from main paper, we adjust the classifier-free-guidance (CFG) scale for ReNoise, InfEdit; Pseudo-Guidance (PG) scale for TurboEdit; and Disentangled Prompt Guidance (DPG) for \name. In Table 4 from main paper, we only adjust DPG scale except for the ablation on PG. For all the experiments, we roughly align on the editability metrics (Alignment). We show that our method produces more consistent results with similar and even better editability. In other words, our method can raise the consistency with lower cost on the editability and vice versa.

\section{Editing Results with Different NFE}
We report the quantitative results of \name with different NFE in ~\tref{tab:supp_nfe}, which shows that the performance of our method scales with increasing NFE. For all settings, we use the same set of hyperparameters as our main experiment. We observe that the consistency metrics improve as the sampling step increases. Another phenomenon we discover is that the alignment metrics do not show a clear trend with the increasing NFE, which is also observed in InfEdit  Table 2.


\section{Further Ablation Results}
We provide more detailed ablation results for the hyperparameters controlling the Controlnet scale and attention mask threshold as in ~\tref{tab:supp_extra_ablation}. The consistency metrics improve with larger ControlNet scale and mask threshold, while the editability metrics experience the opposite. This is expected because larger ControlNet scale and mask threshold tend to maintain contents from the original image thus improving the edit consistency.

\section{Other Baseline Methods with Mask}
We notice that the attention masking mechanism provides a relatively large improvement in the quantitative metric. To ensure that our method's superior performance is not solely attributed to the attention masking mechanism, we extend a similar masking strategy to the baseline methods for a fair comparison. InfEdit already includes a similar masking operation, so we keep it intact. We refer reader to Section 4.1 of Infedit for more details. We present our quantitative results in ~\tref{tab:supp_mask}. We discover that TurboEdit does not benefit from the masking strategy. This might due to the DDPM-noise inversion, as the nondeterministic DDPM noise injects artifacts during the merging process of the source and target guidance signals. ReNoise benefits from the mask as consistency metrics improve. However, the editability metrics drop accordingly. Overall, ReNoise still cannot reach a competitive performance with attention masking.


\section{User Study}
\begin{figure*}[h]
 \centering
 \begin{minipage}{\textwidth}
\includegraphics[width=\linewidth]{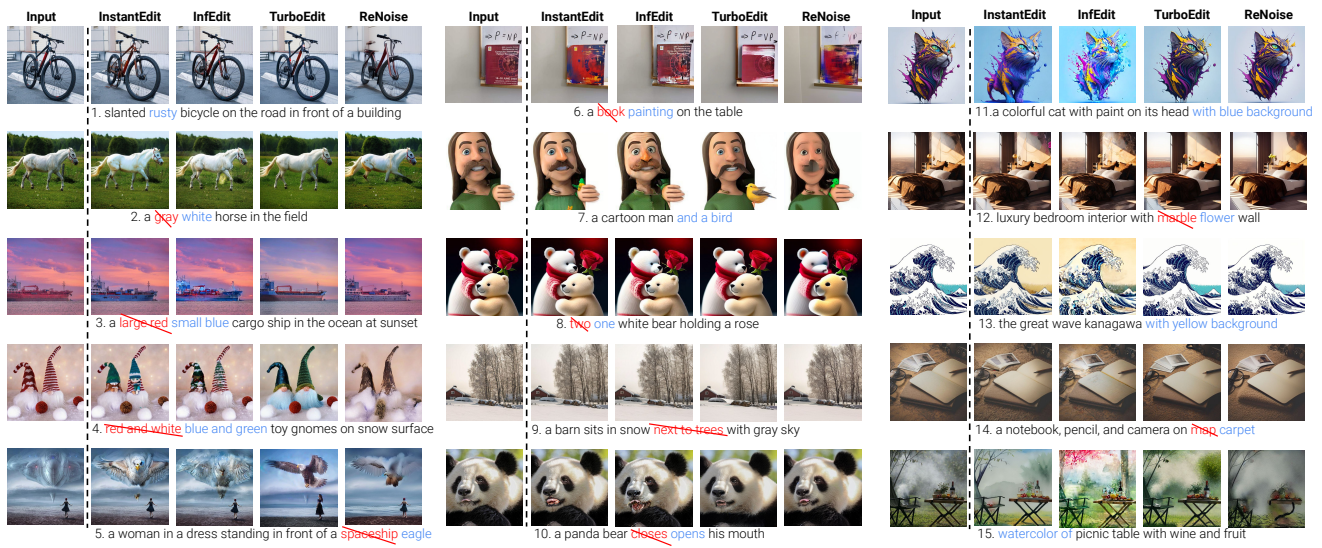}
\caption{ Visualization for randomly selected user study images. 
\label{fig:supp_user_study}
}
\end{minipage}
\end{figure*}

\begin{figure*}[h]
 \centering
 \begin{minipage}{\textwidth}
\includegraphics[width=\linewidth]{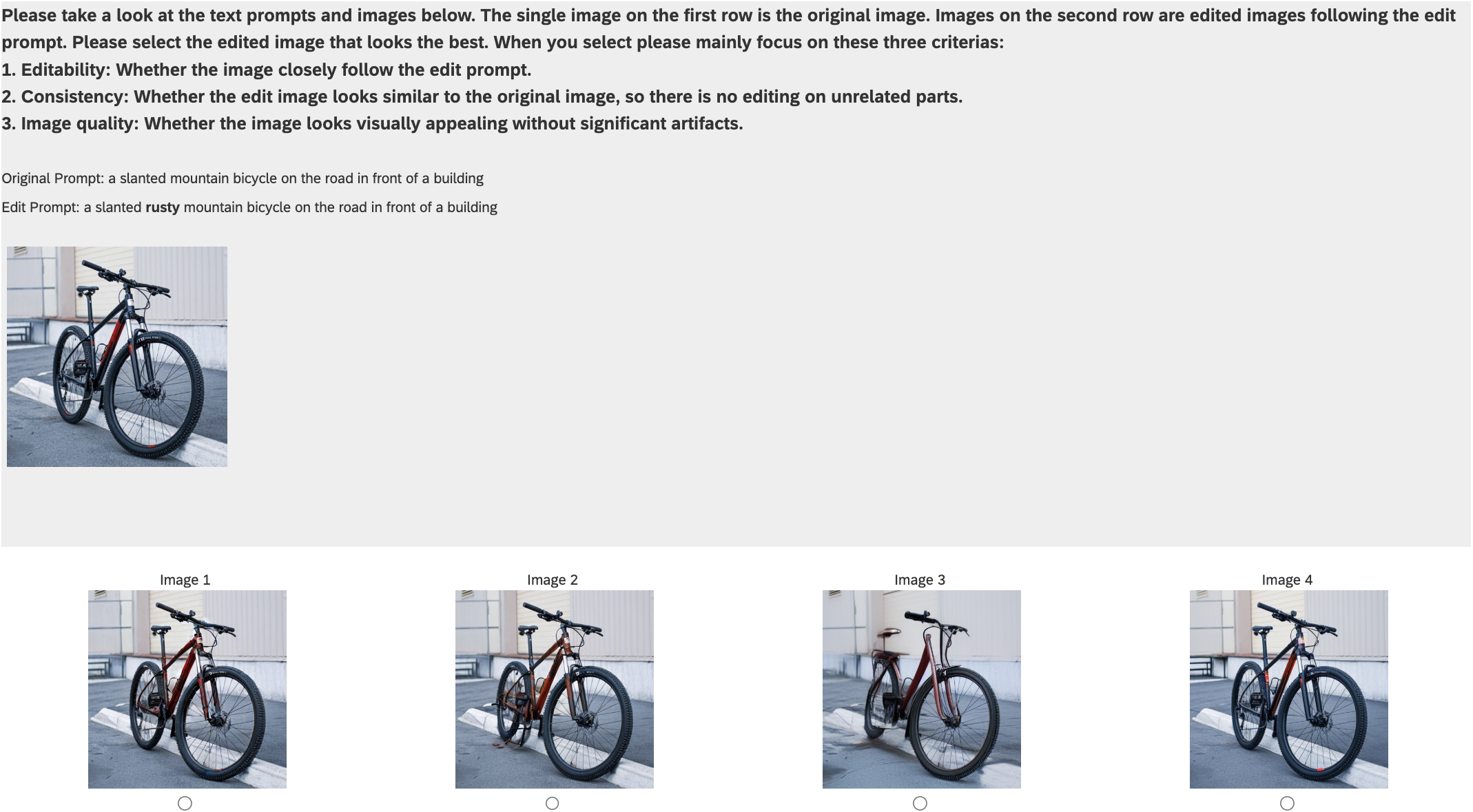}
\caption{ Visualization for our user study interface. We provide general instructions for users to follow when making their decisions. Each method is anonymous and randomly shuffled for  users.
\label{fig:user_study_interface}
}
\end{minipage}
\end{figure*}

\fref{fig:supp_user_study} shows the 15 selected images for user study and \fref{fig:user_study_interface} shows the interface of our user study. We observe that TurboEdit sometimes faces the problem of large structural inconsistency as shown in the case 3 and 5. InfEdit also tends to create noticeable artifacts in 4 steps as shown in case 2, 3, 11. Most of the methods are not able to perform successful editing when large structural change is required like case 7, 8, 9.


\section{Additional Visual Results}
\begin{figure*}[h]
    \centering
        \includegraphics[width=\linewidth]{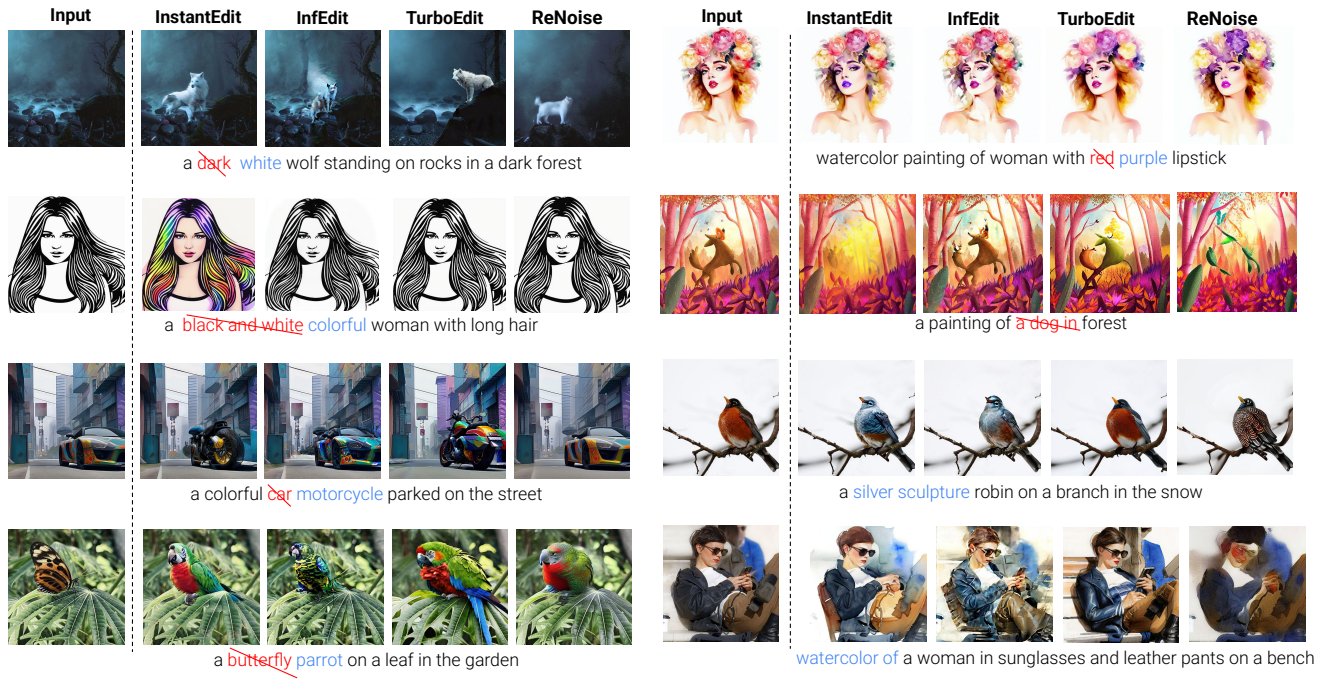}
        \caption{Additional visual comparison with other few-step editing methods.}
        \label{fig:supp_extra_compare}
\end{figure*}

In this section, we show additional comparison results with other baseline few-step editing methods, as shown in Fig \ref{fig:supp_extra_compare}. All the methods perform editing in 4 steps.



\begin{table*}[h!]
    \centering
    \setlength{\tabcolsep}{3pt}
    \renewcommand{\arraystretch}{1.1}
    \label{tab:main_result}
    \begin{tabular}{lccccccc}
       \toprule
        & 
        \textbf{Distance}$_{10^3}^ \downarrow$ & 
        \textbf{PSNR}$\uparrow$ & \textbf{LPIPS}$_{10^3}^\downarrow$ & \textbf{MSE}$_{10^4}^\downarrow$ & \textbf{SSIM}$_{10^2}^\uparrow$ &
        \textbf{Whole}$\uparrow$ & \textbf{Edited}$\uparrow$ \\
        \midrule
        \textbf{DPG: 2.0} & 15.71 & 28.23 & 42.40 & 32.53 & 86.64 & 26.16 & 22.76 \\
        \textbf{DPG: 2.5 (Default)} & 17.14 & 27.96 & 44.39 & 34.94 & 86.44 & 26.28 & 22.82 \\
        \textbf{DPG: 3.0} & 18.71 & 27.70 & 46.39 & 37.39 & 86.22 & 26.33 & 22.87 \\ \hline
    \end{tabular}
    \caption{Quantitative result demonstrating the effect of DPG guidance scale for \name.}
    \label{tab:supp_instant_dpg_scale}
\end{table*}
\begin{table*}[h!]
    \centering
    \setlength{\tabcolsep}{3pt}
    \renewcommand{\arraystretch}{1.1}
    \label{tab:main_result}
    \begin{tabular}{lccccccc}
       \toprule
        & 
        \textbf{Distance}$_{10^3}^ \downarrow$ & 
        \textbf{PSNR}$\uparrow$ & \textbf{LPIPS}$_{10^3}^\downarrow$ & \textbf{MSE}$_{10^4}^\downarrow$ & \textbf{SSIM}$_{10^2}^\uparrow$ &
        \textbf{Whole}$\uparrow$ & \textbf{Edited}$\uparrow$ \\
        \midrule
        \textbf{CFG: 1.8} & 11.85 & 27.78 & 41.55 & 32.48 & 85.82 & 25.31 & 21.88 \\
        \textbf{CFG: 2.3 (Default)} & 16.19 & 26.75 & 50.79 & 42.33 & 84.71 & 25.68 & 22.77 \\
        \textbf{CFG: 2.8} & 28.56 & 24.63 & 73.87 & 71.45 & 82.10 & 26.23 & 22.69 \\ \hline
    \end{tabular}
    \caption{Quantitative result demonstrating the effect of CFG guidance scale for InfEdit. }
    \label{tab:supp_infedit_cfg_scale}
\end{table*}
\begin{table*}[h!]
    \centering
    \setlength{\tabcolsep}{3pt}
    \renewcommand{\arraystretch}{1.1}
    \label{tab:main_result}
    \begin{tabular}{lccccccc}
       \toprule
        & 
        \textbf{Distance}$_{10^3}^ \downarrow$ & 
        \textbf{PSNR}$\uparrow$ & \textbf{LPIPS}$_{10^3}^\downarrow$ & \textbf{MSE}$_{10^4}^\downarrow$ & \textbf{SSIM}$_{10^2}^\uparrow$ &
        \textbf{Whole}$\uparrow$ & \textbf{Edited}$\uparrow$ \\
        \midrule
        \textbf{PG: 0.8} & 14.11 & 25.73 & 66.40 & 44.62 & 83.81 & 25.06 & 21.66 \\
        \textbf{PG: 1.3 (Default)} & 18.57 & 24.59 & 77.53 & 58.48 & 82.64 & 25.70 & 22.30 \\
        \textbf{PG: 1.8} & 35.87 & 21.47 & 117.25 & 117.22 & 78.66 & 26.82 & 23.31 \\ \hline
    \end{tabular}
    \caption{Quantitative result demonstrating the effect of PG guidance scale for TurboEdit. }
    \label{tab:supp_turbo_pg_scale}
\end{table*}

\begin{table*}[h!]
    \centering
    \setlength{\tabcolsep}{3pt}
    \renewcommand{\arraystretch}{1.1}
    \label{tab:main_result}
    \begin{tabular}{lccccccc}
       \toprule
        & 
        \textbf{Distance}$_{10^3}^ \downarrow$ & 
        \textbf{PSNR}$\uparrow$ & \textbf{LPIPS}$_{10^3}^\downarrow$ & \textbf{MSE}$_{10^4}^\downarrow$ & \textbf{SSIM}$_{10^2}^\uparrow$ &
        \textbf{Whole}$\uparrow$ & \textbf{Edited}$\uparrow$ \\
        \midrule
        \textbf{CFG: 5.8} & 19.27 & 25.14 & 80.09 & 50.10 & 82.41 & 25.25 & 21.68 \\
        \textbf{CFG: 6.3 (Default)} & 20.31 & 24.89 & 83.26 & 52.9 & 82.12 & 25.26 & 21.68 \\
        \textbf{CFG: 6.8} & 21.68 & 24.54 & 87.95 & 57.16 & 81.69 & 25.25 & 21.68 \\ \hline
    \end{tabular}
    \caption{Quantitative result demonstrating the effect of CFG guidance scale for ReNoise. }
    \label{tab:supp_renoise_cfg_scale}
\end{table*}
\begin{table*}[h!]
    \centering
    \setlength{\tabcolsep}{3pt}
    \renewcommand{\arraystretch}{1.1}
    \label{tab:main_result}
    \begin{tabular}{lccccccc}
       \toprule
        & 
        \textbf{Distance}$_{10^3}^ \downarrow$ & 
        \textbf{PSNR}$\uparrow$ & \textbf{LPIPS}$_{10^3}^\downarrow$ & \textbf{MSE}$_{10^4}^\downarrow$ & \textbf{SSIM}$_{10^2}^\uparrow$ &
        \textbf{Whole}$\uparrow$ & \textbf{Edited}$\uparrow$ \\
        \midrule
        \textbf{8 NFE} & 17.14 & 27.96 & 44.39 & 34.94 & 86.44 & 26.28 & 22.82 \\
         \textbf{16 NFE}  & 13.64& 29.15 & 36.44 & 26.76 & 87.24 & 26.01 & 22.64 \\ 
       \textbf{24 NFE} & 12.57 & 29.63 & 35.27 & 24.57 & 87.40 & 26.06 & 22.73 \\
         \textbf{32 NFE} & 12.16& 29.69 & 34.95 & 24.36 & 87.49 & 25.96 & 22.69\\  \hline
    \end{tabular}
    \caption{Quantitative results with 8, 16, 24, 32 NFE .}
    \label{tab:supp_nfe}
\end{table*}
\begin{table*}[!b]
    \centering
    \setlength{\tabcolsep}{3pt}
    \renewcommand{\arraystretch}{1.1}
    \label{tab:main_result}
    \begin{tabular}{lccccccc}
       \toprule
        & 
        \textbf{Distance}$_{10^3}^ \downarrow$ & 
        \textbf{PSNR}$\uparrow$ & \textbf{LPIPS}$_{10^3}^\downarrow$ & \textbf{MSE}$_{10^4}^\downarrow$ & \textbf{SSIM}$_{10^2}^\uparrow$ &
        \textbf{Whole}$\uparrow$ & \textbf{Edited}$\uparrow$ \\
        \midrule
        \textbf{TurboEdit} & 25.64 & 24.65 & 109.41 & 54.10 & 80.07 & 25.23 & 21.78 \\
        \textbf{ReNoise} & 26.07 & 25.76 & 65.78 & 53.45 & 83.68 & 24.75 & 21.28 \\ 
        \textbf{InfEdit} &  16.19 & 26.75 & 50.79 & 42.33 & 84.71 & 25.68 & 22.27 \\
        \textbf{\name (Ours)} & 17.14 & 27.96 & 44.39 & 34.94 & 86.44 & 26.28 & 22.82 \\ \hline
    \end{tabular}
    \caption{Quantitative comparison for applying attention mask to all the methods.}
    \label{tab:supp_mask}
\end{table*}

\begin{table*}[!t]
\centering
  \setlength{\tabcolsep}{3pt}
  \renewcommand{\arraystretch}{1.1}
    \begin{tabular}{lccccccc}
      \toprule
       \textbf{Components} & \textbf{Distance}$_{10^3}^\downarrow$ & \textbf{PSNR}$^\uparrow$ & \textbf{LPIPS}$_{10^3}^\downarrow$ 
       & \textbf{MSE}$_{10^4}^\downarrow$ & \textbf{SSIM}$_{10^2}^\uparrow$ 
       & \textbf{Whole}$^\uparrow$ & \textbf{Edited}$^\uparrow$ \\
      \midrule
      ControlNet scale=0.2
        & 18.90 & 27.66 & 47.47 & 38.26 & 86.12 & 26.16 & 22.72 
         \\
      ControlNet scale=0.6 
        & 10.37 & 29.67 & 32.69 & 22.31 & 87.65 & 25.25 & 21.89
          \\
      ControlNet scale=0.8  
        & 8.17 & 30.17 & 29.56 & 19.64 & 88.00 & 24.82 & 21.52
        \\
    \midrule
    Mask threshold=0.2
    & 14.39 & 28.47 & 40.80 & 29.37 & 86.82 & 25.89 & 22.44
     \\
    Mask threshold=0.6
     & 12.88 & 29.20 & 36.08 & 26.75 & 87.24 & 25.61 & 22.22 
      \\
    Mask threshold=0.8
     & 10.00 & 30.29 & 32.03 & 23.27 & 87.68 & 24.99 & 21.56
      \\
      \bottomrule
    \end{tabular}
     \caption{Quantitative results for further ablation 
     experiments}
      \label{tab:supp_extra_ablation}
\end{table*}


\end{document}